\documentclass[letterpaper, 10 pt, journal]{ieeeconf}
\usepackage{amsmath,amsfonts}
\usepackage{algorithmic}
\usepackage{algorithm}
\usepackage{array}
\usepackage[caption=false,font=normalsize,labelfont=sf,textfont=sf]{subfig}
\usepackage{textcomp}
\usepackage{stfloats}
\usepackage{url}
\usepackage{verbatim}
\usepackage{graphicx}
\usepackage{cite}
\usepackage{xcolor}
\usepackage{svg}
\usepackage[normalem]{ulem}
\usepackage{fancyhdr}

\begin{document}

\title{Koala Gripper: Co-designing Robotic Grippers and Data-Capture Devices for Scaling Dexterous Manipulation Learning}

\author{Amar Hajj-Ahmad*,
        Zubin Kremer Guha*,
        Tim Fofonoff,
        Zhi Ern Teoh,
        Ciarán T. O'Neill,
        Ben Thacher,\\
        Igor Fala,
        Vidullan Surendran,
        Murphy Wonsick,
        Peter Whitney,
        David Watkins
        \thanks{* These authors contributed equally to this work.}\vspace{-1.0em}}

\maketitle

%%%%%%%%%%%%%%%%%%%%%%%%%%%%%%%%%%%%%%%%%%%%%%%%%%%%%%%%%%%%%%%%%%%%%%%%

\begin{abstract}
% \vspace{-1.0em}
As the demand for larger manipulation datasets grows, handheld robotic gripper data collection and the associated gripper designs become more vital. Current data collection device designs trend towards matching the morphologies of existing robotic grippers, sacrificing ergonomics and manipulation performance. In this paper, we propose a co-design framework that guides the simultaneous development of both data collection and robotic execution devices by weaving both platform constraints into the design process. Through this workflow, we present the Koala Gripper system, a data capture device and robotic gripper platform that improves dexterity and grasp capability compared to parallel jaw grippers while preserving scalability and ease-of-use. The design introduces a novel force-optimized finger/trigger linkage mechanism with directional reflected mass characteristics, a unique monolithic dual-thumb, and user-centered ergonomic design. The design's actuated robotic fingers are backdrivable, with effective mass on the order of tens of grams. We show that these grippers are capable of secure grasps over a wide range of objects, forceful tool use, and precise singulation. We further validate the platform by deploying it with an end-to-end data collection and policy execution pipeline that highlights its capabilities through learning from demonstration.

\end{abstract}

%%%%%%%%%%%%%%%%%%%%%%%%%%%%%%%%%%%%%%%%%%%%%%%%%%%%%%%%%%%%%%%%%%%%%%%%

\section{Introduction}
\label{sec:intro}
Dexterous manipulation and object handling in unstructured environments continue to present challenges in the robotics field. Historically, there has been success using task-specific grippers designed for assembly line automation. Deploying classical control strategies on highly stiff and repeatable systems works well for execution in structured environments \cite{Siciliano2008SpringerRobotics}. However, the aspirations of the robotics field are moving towards solving manipulation in unconstrained and highly unstructured environments, similar to those where humans operate. As a result, learning-based policies have emerged as a powerful tool for controlling robotic grippers in these conditions where more adaptive and generalizable skills are required \cite{Andrychowicz2020LearningManipulation, Padalkar2023OpenModels}. However, compared to the internet-scale data available for language models and vision models, robotic manipulation has relatively little publicly available data. Thus, data collection and generation has become a key area of focus for solving manipulation.

\begin{figure}[!h]
    \centering
    \includegraphics[width=0.95\linewidth]{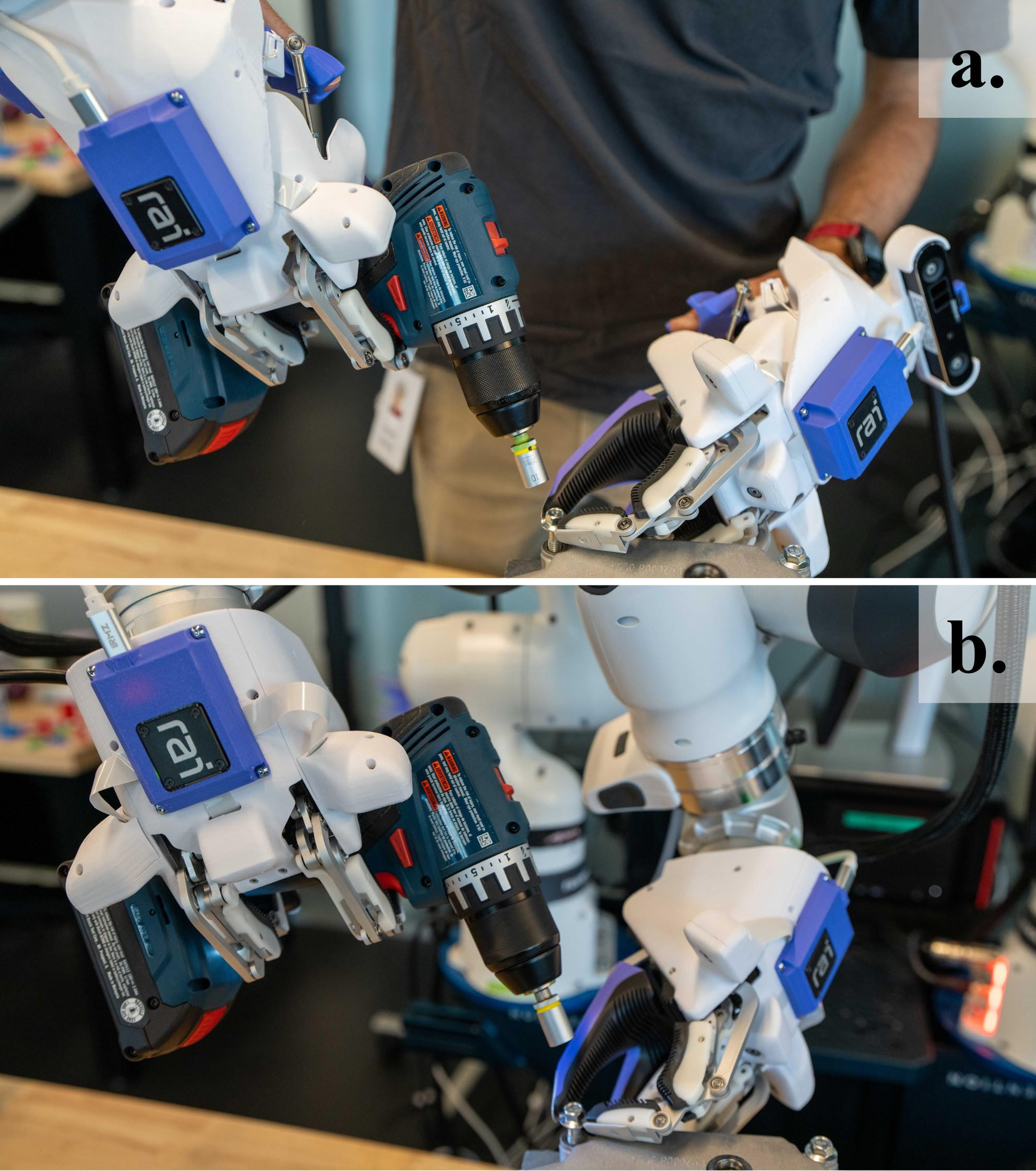}
    \caption{The co-designed Koala Data Capture Device (a) and Robot Device (b) demonstrate a bolt fastening task using a standard drill.}
    \label{Hero}
\end{figure}

Currently, the prevailing methods for data collection include simulation, teleoperation, and handheld data capture: 

\textbf{Simulation} methods are powerful tools for enabling robotic manipulation performance, both through large-scale synthetic dataset generation \cite{Mandlekar2023MimicGen:Demonstrations,Jiang2024DexMimicGen:Learning}, and reinforcement learning \cite{Andrychowicz2020LearningManipulation}. These strategies sidestep the need for physical robot hardware and data collector time. However, realistic scene and asset creation can be laborious, and the sim-to-real gap a significant obstacle, especially for contact-rich manipulation tasks. 

\textbf{Teleoperation} bypasses the sim-to-real gap by directly interacting with the real world through robot hardware \cite{Zhao2023LearningHardware,Khazatsky2024Droid:Dataset,Wu2024GELLO:Manipulators,Qin2023AnyTeleop:System}, resulting in a minimal embodiment gap during policy execution on the same robot the data was collected. However, large-scale teleoperation data collection can be expensive, requiring multiple robotic systems. It also requires significant operator skill to overcome the lack of proprioception and haptic feedback for most teleoperation systems.

\textbf{Handheld data capture} places the manipulation device in the operator's hand, making the system extremely portable and far less complex and costly when compared with teleoperation \cite{Chi2024UniversalRobots,Liu2025FastUMI:Dataset,Fang2025DEXOP:Manipulation,Wu2025MagiClaw:Gap}. While handheld data capture suffers a larger training-to-execution gap than teleoperation, its strength lies in a smaller human-to-robot gap, allowing it to leverage human intuition in solving and executing complex manipulation tasks.

Handheld data capture methods have the potential to enable large-scale real-world robotics datasets. However, current state-of-the-art handheld data capture devices have been limited to parallel jaw grippers \cite{Chi2024UniversalRobots,Liu2025FastUMI-100K:Dataset,Liu2025ViTaMIn:Interface,Seo2025LEGATO:Tool}, capture devices tied to pre-designed robot devices \cite{Xu2025DexUMI:Manipulation, Zhu2026DexEXO:Learning, Zhu2026WHED:Collection}, and implementations with the human hand within the gripper workspace \cite{Wei2024ALearning,Xu2025DexUMI:Manipulation,TechnologyCompany}, limiting manipulation capabilities. Additionally, exoskeleton inspired data collection devices identify their limitations in ergonomics and generalization across users \cite{Xu2025DexUMI:Manipulation, Romero2024EyeSightActuation, Zhu2026WHED:Collection, Zhu2026DexEXO:Learning, Chao2025Exo-ViHa:Learning, Du2025MILE:Manipulation, Tao2025DexWild:Policies}. Therefore, we introduce a co-design framework for developing data capture device (CD) and robot device (RD) pairs to unlock more gripper performance. In this framework, both systems are designed simultaneously to balance their constraints and capabilities, avoiding design decisions that disadvantage one over the other. This process is used to design the Koala gripper architecture, a pair of grippers featuring a bio-inspired trigger mechanism coupled with novel underactuated linkage-based fingers and a grasp-mode-switching dual-thumb (as seen in Fig.~\ref{Hero}). Its ergonomic features, including support surfaces, three standard sizes, and cushioning grip padding, are designed to reduce operator fatigue. This work characterizes the performance of the CD and RD in terms of force output and grasp execution. The work also highlights the underactuated linkage mechanism's selective low reflected inertial properties and the utility of an intentionally designed back-drivable system. The design process and resulting devices are validated through imitation-learning-based policy execution that requires human-collected data and results in successful grasps and task completion beyond the capabilities of parallel-jaw grippers.

%===================================================================================

\section{Design Methodology}
\label{Design Methodology}

The co-design process begins by defining the task requirements of the system, which are the set of capabilities necessary for both devices to effectively complete a desired set of tasks. Some examples of these requirements are a maximum object grasp size, executing a specific grasp type, and a multi-modal sensing package for detecting gripper state. It is also important to define the desired workspace for the gripper to shape how it interacts with the environment and the user. These targets serve to start the process of designing key common elements between both devices. Subsequently, constraints and performance targets specific to the handheld data capture device (CD) and robotic gripper device (RD) must be defined. 

\textbf{For the CD}, operators introduce many constraints around the user interface of the device. Those most notable include user ergonomics, mechanical advantage, and degrees of freedom (DoF). Good ergonomics are critical to minimize user fatigue when capturing data for extended periods of time and support the manipulation of heavy or cumbersome objects. Sufficient mechanical advantage is important for producing forces for grasping, which is especially relevant when the output stroke of the CD fingers is greater than the physical inputs of the device, deleveraging the user. Finally, the number and modality of DoFs controllable by the user must be intentionally managed to balance user mental load with gripper performance while allowing them to securely hold the device. 

\textbf{For the RD}, most constraints come from actuators and actuator packaging. For instance, low-inertia actuators are key to allowing the RD to react more similarly to the relatively low-inertia CD, enabling safer and compliant contact with the environment. In turn, actuator packaging constraints also influence the layout and overall size of the gripper.

\begin{figure}[!t]
    \centering
    \includegraphics[width=\linewidth]{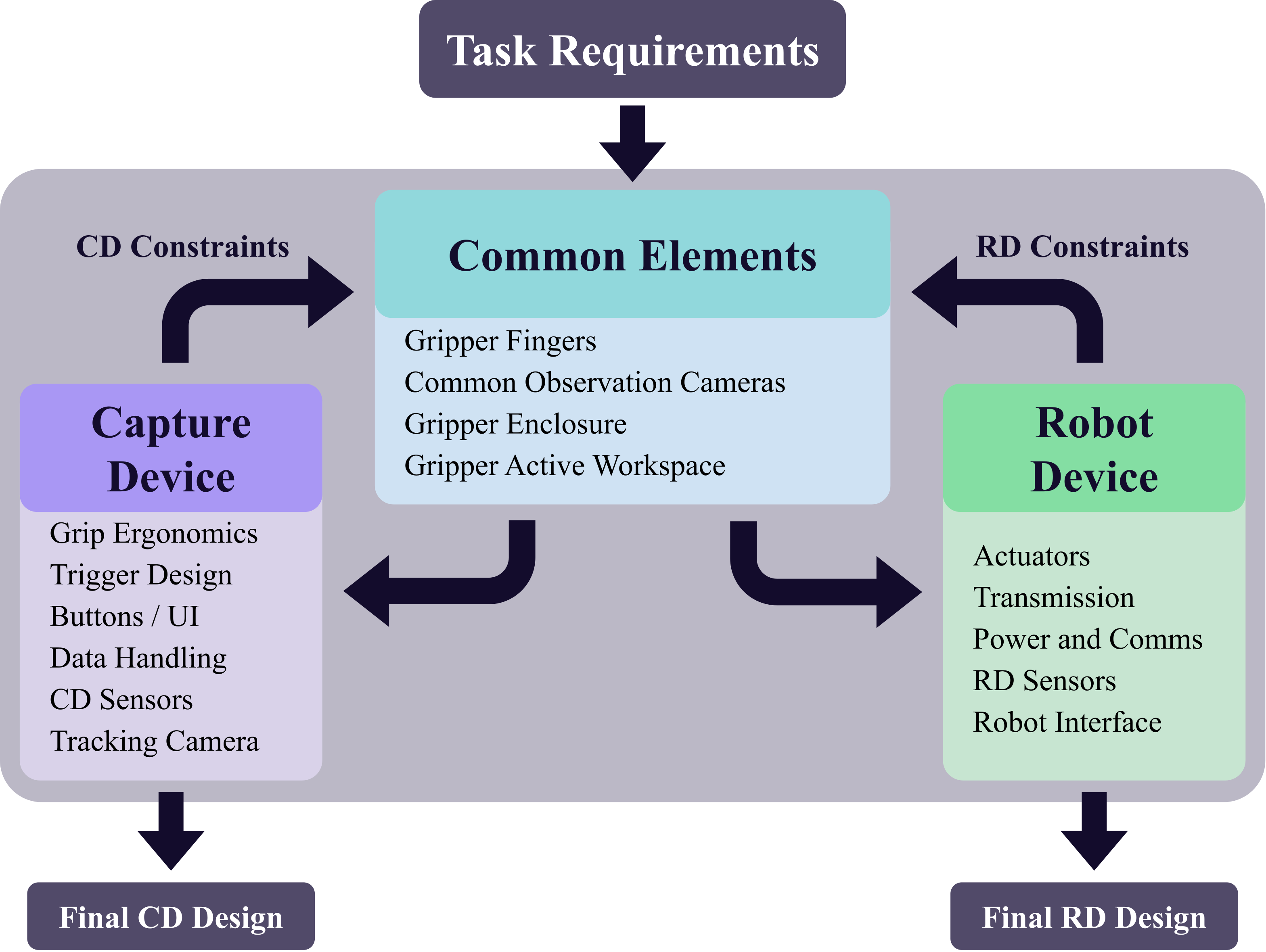}
    \caption{Task requirements, along with constraints and design targets from the CD and RD, feed common features and elements. These are then propagated back to each device, creating iterative cycles throughout the co-design process.}
    \label{design_block}
\end{figure}

As the design for the system develops, these device specific constraints and targets interact with each other and the common system elements, creating the feedback loop shown in Fig.~\ref{design_block}. In this way, the common elements may evolve as details of the CD and RD necessitate it. This workflow pairs well with a master modeling approach where both devices directly reference the same elements, ensuring that they stay aligned to common features.

%===================================================================================

\section{Design Implementation}
\label{Design Implementation}

This section describes the co-design process for designing the Koala gripper system, highlighting the key characteristics of the design.  A key goal was designing a device that allows a robotic system to use tools and complete tasks intended for a human operator.

\begin{figure}[!b]
    \centering
    \includegraphics[width=\linewidth]{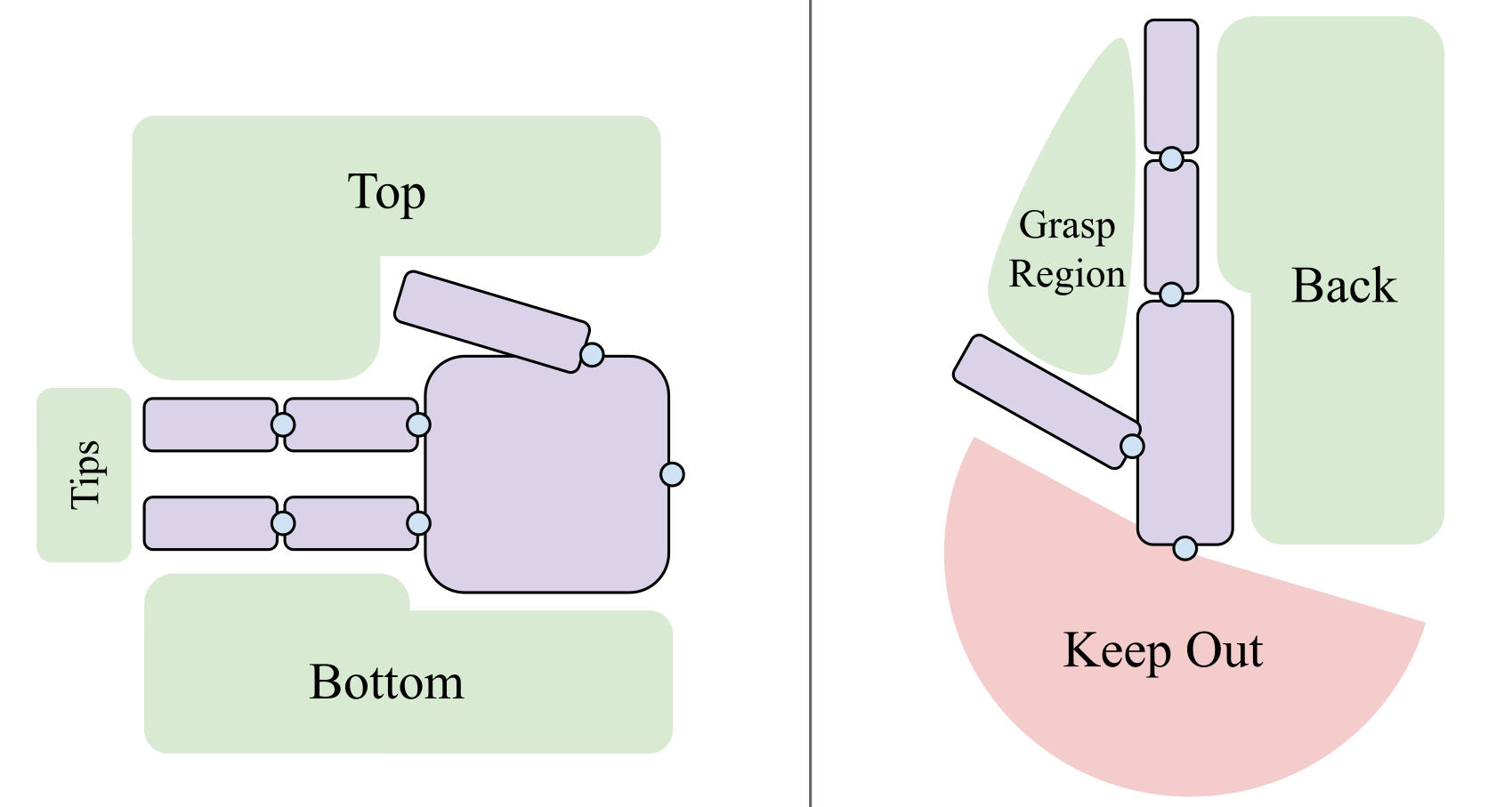}
    \caption{The space around a robotic gripper can be segmented into several key active workspace areas that define how the gripper interacts with the environment and the associated design implications.}
    \label{active_workspace}
    \vspace{-2mm}
\end{figure}

\subsection{Gripper Active Workspace}

The co-design process begins by defining key areas of the gripper workspace, as defined on a generic gripper in Fig.~\ref{active_workspace}, for informing gripper form factor decisions. Keeping the "bottom" region clear allows for table-top surface picks, and the boundaries for the "top" and "back" regions should comply with \textit{access opening dimensions} as defined in \cite{DepartmentsandAgenciesoftheDepartmentofDefense2020DEPARTMENTENGINEERING}. Features on the device such as cameras and fingers must be positioned so as not to interfere with generic shapes of standard hand tools. Keeping the "tips" region clear allows for button pushing or probing. A "keep-out" region ensures that the CD user's range of motion is not limited. Keeping the device under specific dimensions coupled with maintaining surfaces for eventual sensor integration reinforces the design choice to restrict the active workspace from the user's hand.

\begin{figure*}[!t]
    \centering
    \includegraphics[width=\linewidth]{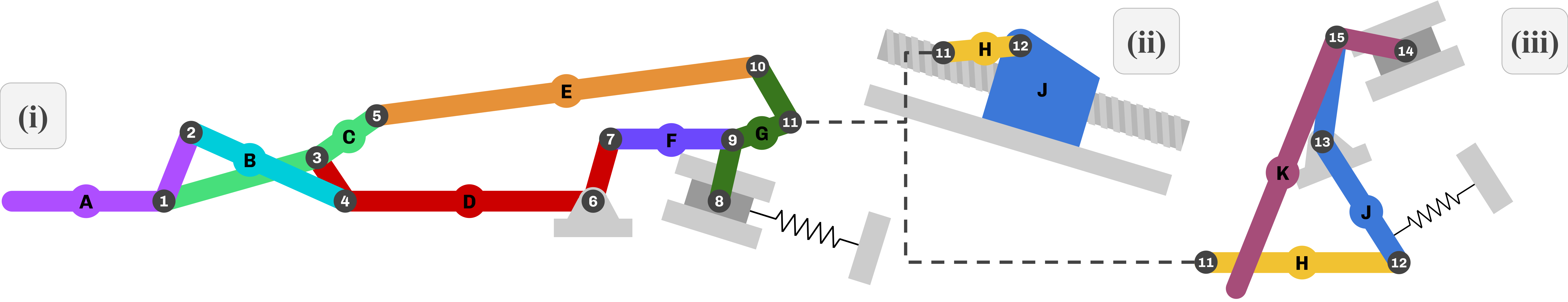}
    \caption{The finger linkage mechanism integrates the common finger linkage (i) with the RD actuation mechanism (ii) and the CD trigger mechanism (iii)}
    \label{links}
\end{figure*}

\subsection{Gripper Morphology and Grasp Taxonomies}

Minimizing physical and mental loads on the user are integral goals for a data capture device, which is supported by limiting the user-controllable DoFs of the gripper. However, the dexterous behaviors targeted by the Koala gripper design include things like forceful tool use with torsional stability, trigger actuation and some reorientation. These tasks often require multiple DoFs from human hands, driving this gripper design towards non-anthropomorphic morphologies and grasp strategies that can succeed at these tasks with less DoFs.

For instance, the same success metrics of lateral grasp types requiring thumb adduction \cite{Feix2015TheTypes} can be achieved with powerful pinch grasps, eliminating the need for an abducting thumb and  reducing DoFs. However, to achieve both power and pinch grasps with the same finger, it must still be possible for the finger to meet the thumb for fingertip grasps and bypass it for smaller wrap grasps. This can be implemented with low DoF fingers in several ways, including pairing 2-DoF fingers with a static thumb \cite{TechnologyCompany} or matching 1-DoF fingers with a reconfigurable thumb. Thus, a simple pivoting thumb achieves this behavior while reducing user-actuated DoFs. 

At the finger level, grasps targeting trigger operated tools call for at least two independently-actuated non-opposed fingers. Preshaping in these fingers helps achieve hooking grasps and promotes enveloping grasps that prevent object ejection. Additionally, underactuation allows for conforming to irregular objects, more secure wrap grasps on objects of different handle sizes, and trigger actuation on a variety of tools such as drills and spray bottles. 

Extending the 1-DoF thumb to oppose both independently actuated fingers establishes moment balance stabilizing forceful power grasps, something that parallel jaw grippers fail to provide. This concept of a monolithic "dual thumb", which inspired the Koala gripper's name, also achieves the goals of prismatic grasps \cite{Feix2015TheTypes}, facilitating writing, fine tool stabilizing, multi-object singulation, table-top pinches, and stable corner/palmar grasps. Ultimately, the Koala morphology constitutes two 1-DoF preshaping underactuated fingers opposing a 1-DoF pivoting dual thumb, totaling three controllable DoFs.

\subsection{Finger Mechanisms}

Co-designing for a handheld and motorized gripper introduces a unique set of constraints to the finger design space. For instance, serial, fully-actuated finger designs are difficult to integrate into capture devices because of their high DoF count, while tendon-driven fingers require complex assembly and tensioning procedures that make them difficult to reliably manufacture and service at scale. This, combined with the preshaping behavior of the chosen hand morphology, led to the choice to develop a linkage-based finger mechanism.

\textbf{Finger Linkage} 
The Koala finger is comprised of a 9-bar linkage, with one underactuated and one actuated DoF. The design was adapted from an existing 7-bar mechanism \cite{Wang2025ForceFeedback}, integrated with a crossed 4-bar fingertip and modified at link (F) (Fig.~\ref{links}). The mechanism has a grounded underactuation spring and a favorable lack of incursion into the "palm" area.

In Fig.~\ref{links}, distal link (A) is coupled to proximal link (D) via a crossed 4-bar (B \& C) which curls the finger as it closes. The curling behavior is compounded by the length differential between links (E) and (F), which drive the crossed 4-bar and proximal link, respectively.
The underactuated joint (8) is connected to a prismatic stage located proximal of the finger, placing it inside the final device enclosure. When proximal link (D) makes contact or meets a hard stop, underactuation allows distal link (A) to continue closing. The finger is designed to be driven by a push-rod attached to link (G) at joint 11.

\textbf{Thumb Linkage} The dual monolithic thumb, moves between fingertip and power grasp positions by pivoting about a joint, making it a simple 1-DoF system. Simple 4-bar mechanisms actuate the thumb in both the CD and RD. The initial and fully extended thumb positions are used for pinch and power grasps, respectively, reducing the positioning precision and cognitive load on the user. The driving linkage leverages a singularity for pinch and a hard stop for power to separate the user from the load path of the thumb.

\subsection{Finger Actuation}

\begin{figure}[!t]
    \centering
    \includegraphics[width=0.8\linewidth]{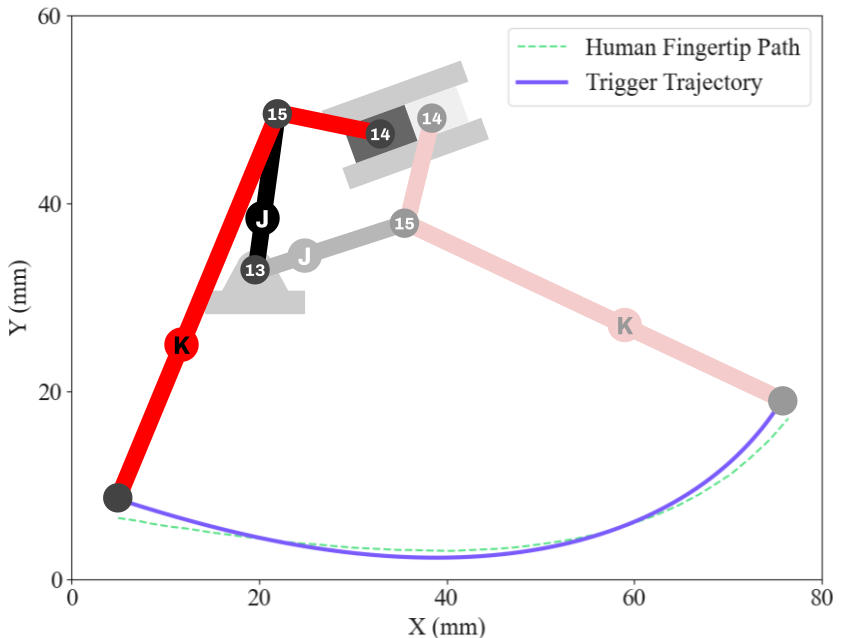}
    \caption{The path traced by the proposed trigger mechanism closely mimics natural fingertip closing paths \cite{Lu2024FingerVision}, improving energy harvesting performance.}
    \label{trigger}
\end{figure}

\textbf{CD Actuation} Handheld capture devices that rely on mechanical power are limited by user comfort and strength. It becomes imperative to efficiently and ergonomically harvest energy from the user. One simple way to improve energy harvesting is lengthening the trigger stroke, increasing total work per stroke. However, longer trigger strokes can be uncomfortable or even unusable when applied to a linear trigger. To overcome this limitation, Koala uses a 4-bar linkage trigger mechanism that closely mimics the natural closing motion of a human fingertip \cite{Lu2024FingerVision} (Fig.~\ref{trigger}). This mechanism uses a prismatic joint at joint (14), allowing the trigger to trace a lower curvature path than with a single revolute joint, streamlining packaging and improving mechanical advantage.

\begin{figure}[!t]
    \centering
    \includegraphics[width=0.95\linewidth]{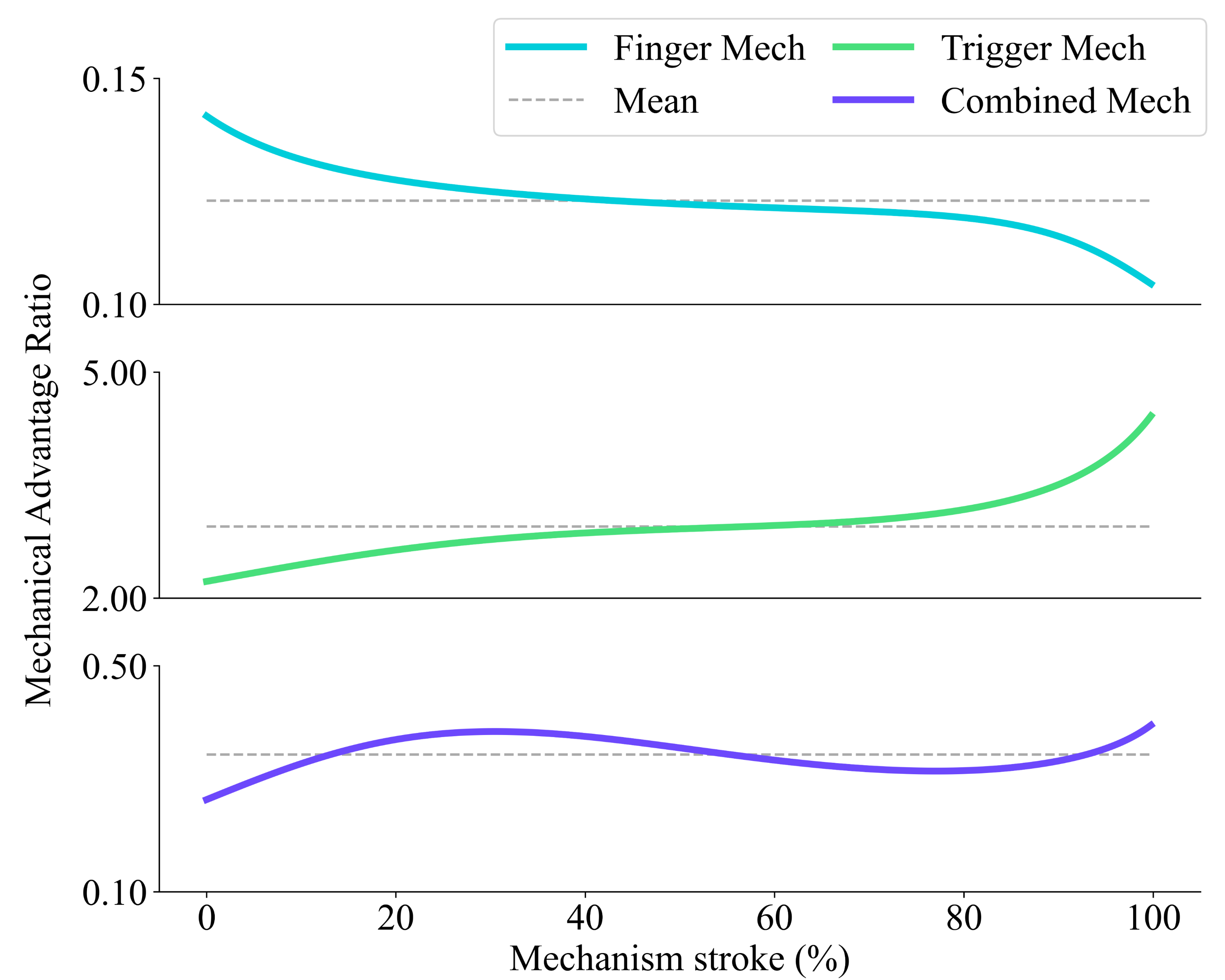}
    \caption{Mechanical advantage of the finger (top), trigger (middle), and combined (bottom) mechanism as configured in the CD. Augmenting the two mechanisms creates consistent mechanical advantage over the full stroke.}
    \label{mech_adv}
\end{figure}

This trigger linkage, when used as an input to the finger linkage (Fig.~\ref{links}\,(i,iii)), serves to counteract deleveraging as the finger approaches end-of-stroke singularities. The mechanical advantage curves from each sub-linkage counteract each other, resulting in a flatter curve for the full finger assembly (Fig.~\ref{mech_adv}). 

\textbf{RD Actuation} was designed to be packaged compactly within the gripper while maintaining the force output and low inertia characteristics of the handheld CD. Actuators interface with shared finger parts to reduce unique part count. Both planetary gear reductions and ballscrews were considered for their high efficiency and backdrivability, which aid in replicating the low inertia characteristics of the CD as characterized and discussed in Section \ref{characterization}. A 4\,mm pitch ballscrew and linear guide configuration paired with a TQ ILM25x08 frameless motor was chosen for the finger actuation, having preferable energy-per-cycle (10\,J/cycle) and effective packaging. The actuator drives a 4-bar linkage (Fig.~\ref{links} (i)) designed to match the peak output force of a human operator, assuming a 100\,N maximum applied force per trigger. An 18.8:1 planetary reduction paired with the same motor was chosen for the thumb actuation to match the higher torque requirement of the CD thumb actuation.

\subsection{Active Surfaces}

\begin{figure}[!h]
    \centering
    \includegraphics[width=\linewidth]{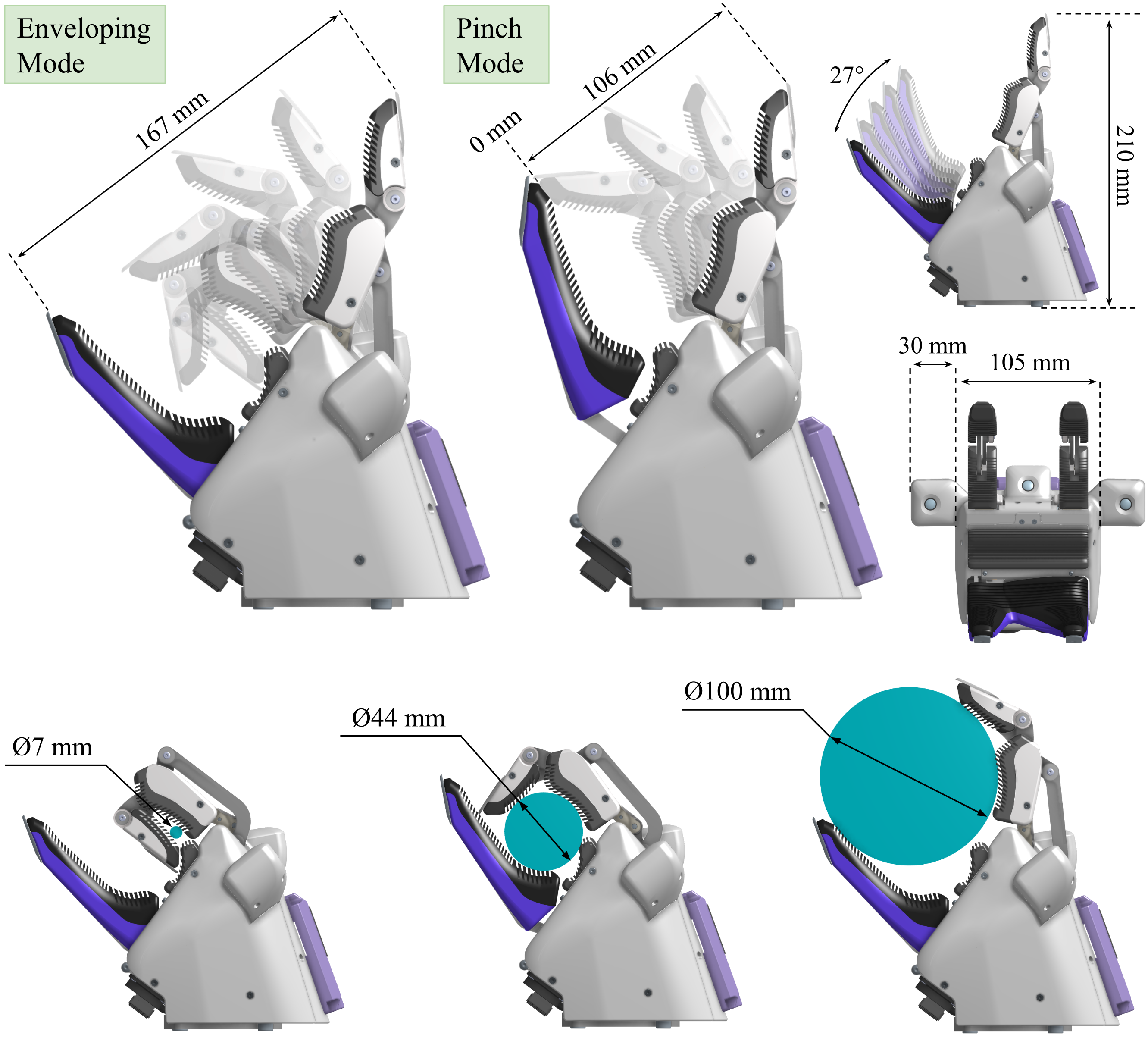}
    \caption{The full workspace of the Koala gripper spans from 0mm pinch grasps to power grasps covering a 7\,mm\,--\,100\,mm diameter range. Pad geometry targets power grasps of a cylindrical 44\,mm diameter standard handle \cite{HenryDreyfuss1955DesigningPeople}.}
    \label{workspace}
\end{figure}

Designing the contoured shapes of the device's thumb, fingers, and palm was heavily influenced by the hand data and tool design metrics from \cite{HenryDreyfuss1955DesigningPeople}. The thumb has an asymmetric shape which closely resembles the natural opposing pad of a human thumb, as documented in \cite{HenryDreyfuss1955DesigningPeople}, for tool use. The finger pads are meant to achieve pinch grasps and a range of power grasps. The power grasps cover large objects 100\,mm wide, wrap grasps of 44\,mm handles (the average size of handles designed for humans \cite{HenryDreyfuss1955DesigningPeople}), and small tool power grasps of 7\,mm, comparable to a standard pencil (Fig.~\ref{workspace}). 

We empirically determined that pads with 2-3\,mm of compliance were required to grasp cylinders of this range of sizes. Cantilevered slanted features added to the pad surface achieve such compliance without reducing shore hardness, which would sacrifice material durability. The features deflect when compressed by a grasped object, providing a jamming force and improving contact area with irregular surfaces. A Shore 30A polyurethane was chosen for the pad material to balance durability, compliance, and friction, the last of which is vital for achieving secure palmar grasps of large and wide objects. Each pad is vacuum molded onto a rigid ABS-like structure that is then fastened to the metal finger links.

Rigid metal fingernails were also incorporated into the finger and thumb designs in order to achieve precise singulation tasks. These are critical for picking up thin flat objects off flat surfaces, singulating small objects, turning pages, and a variety of tasks that require a zero thickness pick.

\subsection{Handheld Device Ergonomics}

When designing the handheld version of the gripper, it was imperative to make the device accessible and usable for a wide range of users in both unimanual and bimanual configurations. To this end, measurements were collected through a survey of 30 potential users to understand the distribution of hand sizes and finger dimensions. The design goals of the device prioritized wrist and grip support, usability across different hand sizes, and unassisted unimanual and bimanual equipping.

\begin{figure}[!t]
    \centering
    \includegraphics[width=\linewidth]{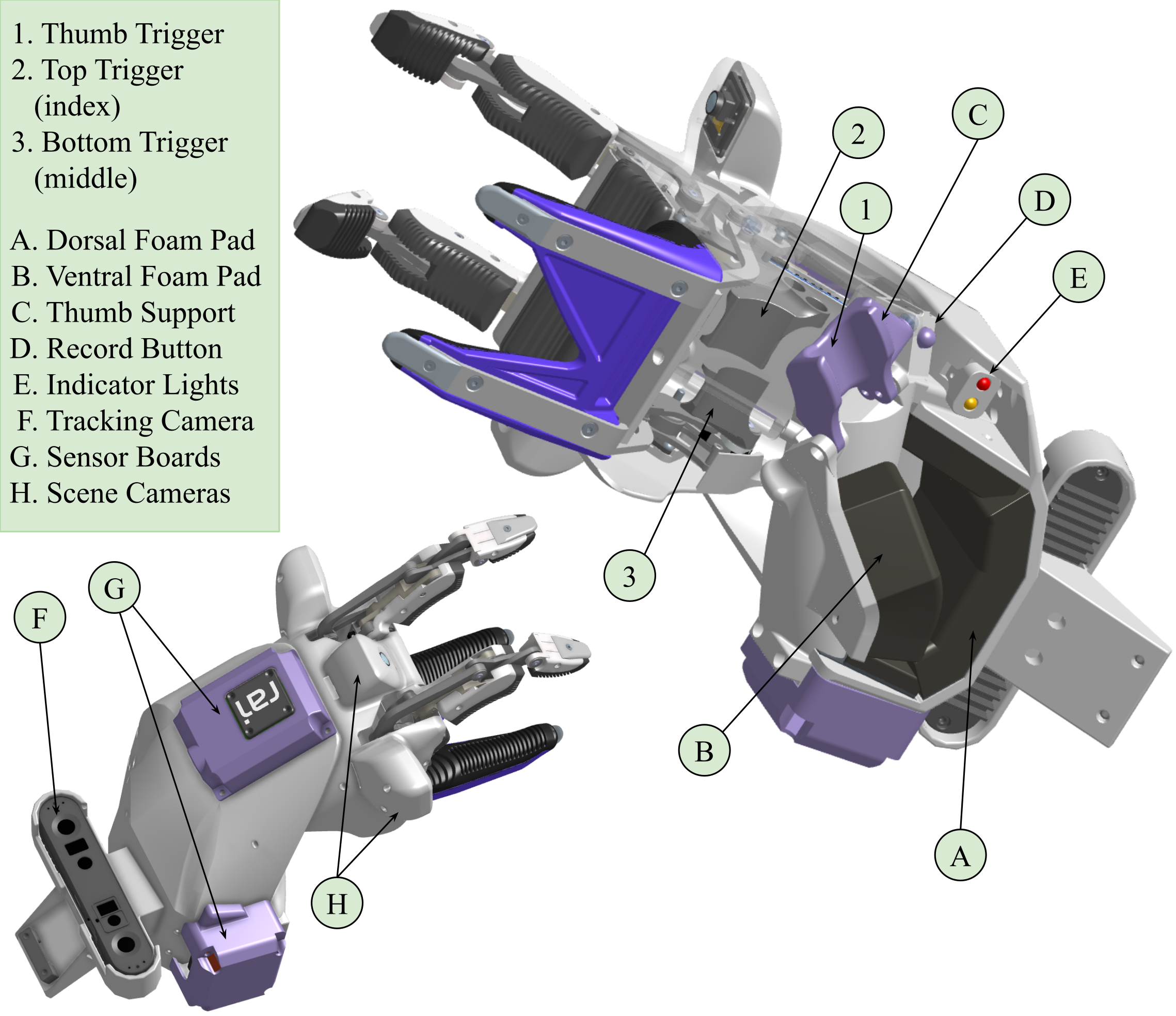}
    \caption{Triggers for the thumb (1), index finger (2), and middle finger (3) map human actuation to the gripper fingers and thumb. Stabilizing ergonomic features include a dorsal cushion (A), a ventral cushion (B), and a thumb-trigger side load support (C).}
    \label{ergo}
\end{figure}

Surveyed hand dimensions informed the design of three distinct grip sizes defined by three parameters: the distance of the device grip to the device fingertips, the distance from the device grip to three actuation triggers, and the thickness of the foam inserts securing the user's hand. The triggers themselves were designed with organic curved surfaces that followed the contour of users' fingers as seen in Fig.~\ref{ergo} (1,2,3).

The trigger for the thumb DoF is canted to the side to promote engagement of the user's palm with the grip. This, coupled with the snug fit from the dorsal and ventral foam inserts (Fig.~\ref{ergo} (A, B)) relieve the user from gripping the device using their two weakest fingers, the ring and little.

\subsection{Sensor System}

For policy training and execution, a sensing system was set up to accurately measures the state of the gripper. The state of the gripper is defined as its 6-DoF pose, the position of the three controllable DoFs, and camera views of the gripper-object interaction and the surrounding scene. The underactuated DoFs of the fingers were omitted from the state because they are not directly controllable.

Although the 6-DoF pose and controllable DoFs are directly measured on the RD by the robotic arm and gripper actuators, they must be explicitly measured on the CD. Magnetic encoders are used to measure finger and thumb linkage positions via a custom multi-sensor interface board, which are then converted to RD actuator position via a lookup table. A 6-DoF pose is measured with an Xvisio Seersense DS80, a visual odometry-based tracking camera.

\begin{figure}[!b]
    \centering
    \includegraphics[width=0.9\linewidth]{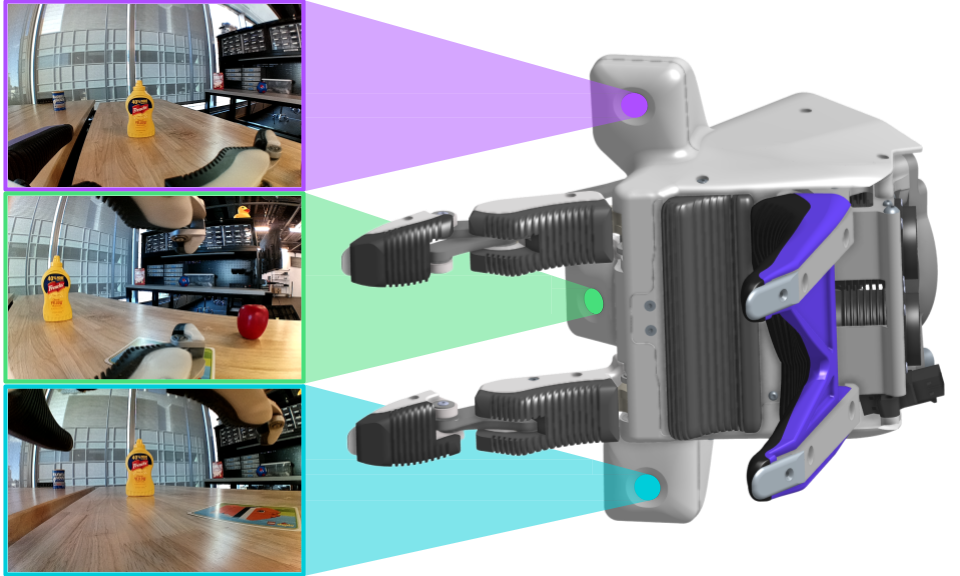}
    \caption{Three wide-angle cameras are positioned on the gripper to observe the gripper fingers and the scene from multiple angles with minimal occlusions.}
    \label{cameras}
\end{figure}

Since gripper-mounted cameras are designed to be the primary visual input of the system, the grasped object and scene must be visible as much as possible, regardless of the object and grasp type. Additionally, key manipulation points like the gripper fingertips needed to be visible from multiple camera views to aid in depth perception for fine manipulation. The final camera array, as shown in Fig.~\ref{cameras}, features two side-mounted and one center-mounted wide-angle Luxonis OV9782 global shutter cameras.

\subsection{Data Collection}

The co-designed nature of the Koala gripper enables two complementary data collection modes: handheld in-the-wild collection with the CD, and teleoperated collection with the RD using the CD as an input device. The handheld mode maximizes task and scene diversity at low marginal cost per demonstration, while the teleoperated mode provides smaller volumes of higher-fidelity on-robot data while adhering to robot's real-world dynamics used to execute the policy. Notably, the shared architecture of the CD and RD makes their state and action streams compatible without re-targeting. 

In both modes, status LEDs and a button on the CD allow the user to start, stop, and monitor the status of the data collection session without setting down the CD. Live camera streams displayed on the host computer aid the collector in ensuring that relevant objects remain in view of the gripper throughout the demonstration.

%%%%%%%%%%%%%%%%%%%%%%%%%%%%%%%%%%%%%%%%%%%%%%%%%%%%%%%%%%%%%%%%%%%%%%%%

\section{Results}
\label{Testing}

\subsection{System Characterization}
\label{characterization}

\textbf{Simulation}
The CD fingers are highly backdrivable because forces are transmitted via low-friction bearings and low-stiffness springs to the user's finger. Consequentially, it is important for the RD to also be backdrivable to exhibit similar compliance to the CD during contact-rich tasks. This can be characterized by calculating the effective mass of the RD fingers with the actuators in the loop. Since linkages can exhibit singularities in specific directions, $m_{eff}$ was calculated tangent to the trajectory of linkage fingertip. 

\begin{figure}[!h]
    \centering
    \includegraphics[width=\linewidth]{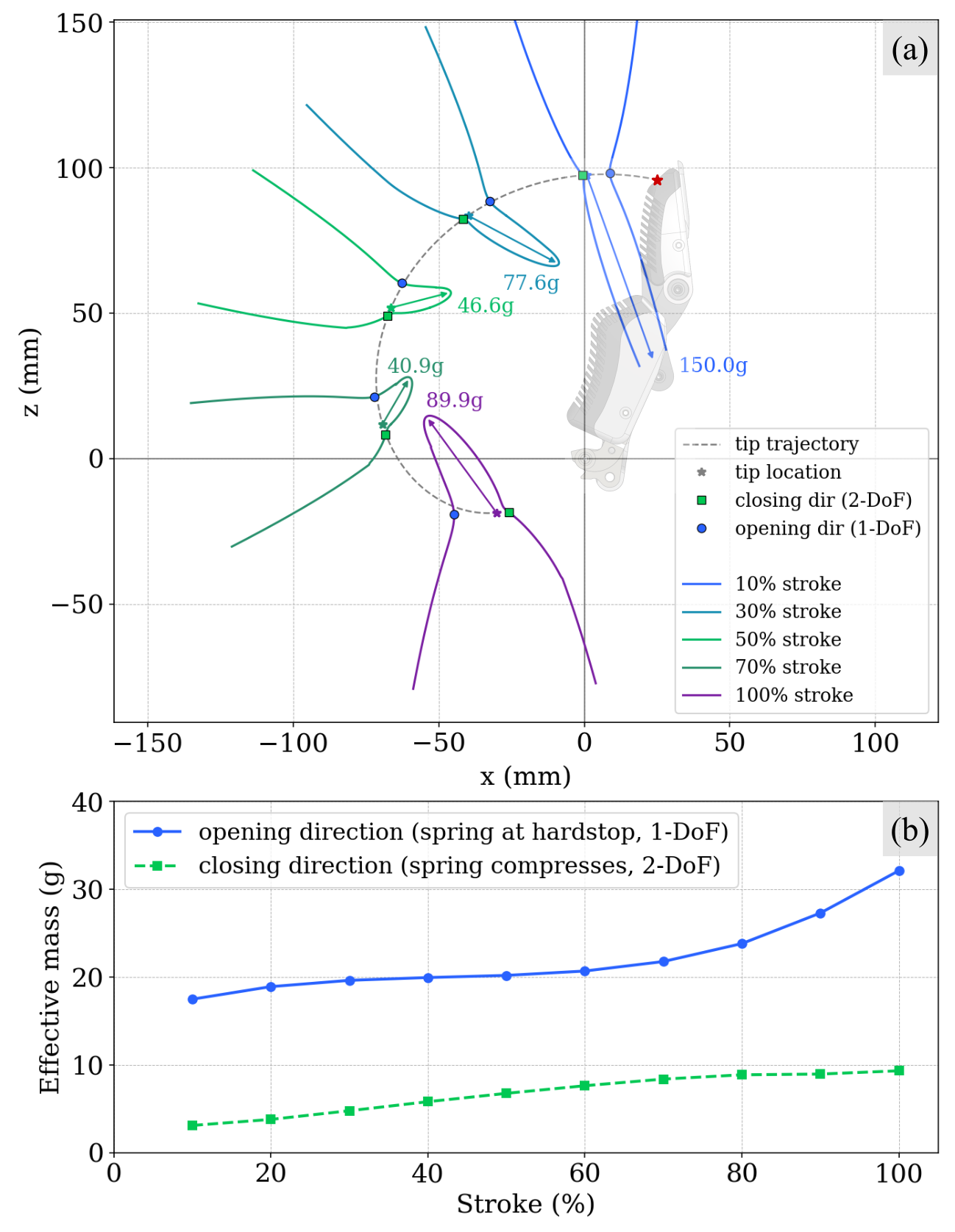}
    \caption{Belted ellipsoid (a) and path-aligned (b) representations of finger effective mass over its complete stroke.}
    \label{m_eff}
\end{figure}

The system was modeled in MuJoCo using equality constraints to define the closed-loop tree. With the mechanism being planar by design, the analysis is done in the x-z plane as depicted by the overlay in Fig.~\ref{m_eff}. Since a single finger is a complex closed kinematic chain, the inertial characterization is done with a finite difference approach, leveraging MuJoCo's physics engine to compute the system's reduced Jacobian. The reduced Jacobian takes into account the velocities induced by the active (finger actuator) and passive (underactuation spring) DoFs of the system. The resulting matrix is:

\begin{equation}
J_{red} =
\begin{bmatrix}
    v_{tip_{1_x}} & v_{tip_{2_x}}\\
    v_{tip_{1_y}} & v_{tip_{2_y}}
\end{bmatrix}
\end{equation}

Where $J_{red}$ is a 2x2 matrix mapping the velocities induced at the tip by the actuator and spring DoFs as $v_{tip_1}$ and $v_{tip_2}$ respectively. 

The reduced mass matrix of the system is found similarly by projecting the system's mass matrix $M_{full}$ along the constraint-consistent velocity vectors $\dot{q_1}$ and $\dot{q_2}$, which are rank 10 vectors relating the 10 joint motions to the actuator and spring perturbations, respectively.

\begin{equation}
M_{red} = 
\begin{bmatrix}
    m_{11} & m_{12}\\
    m_{21} & m_{22}
\end{bmatrix} \\
\end{equation}

\begin{equation}
m_{ij} = \dot{q}_i^T * M_{full} * \dot{q}_j
\end{equation}

Using $J_{red}$ and $M_{red}$, the operational space inertia matrix at the finger tip $\lambda_{tip}$ is represented as follows:

\begin{equation}
    \lambda_{tip} = (J^T_{red}*M^{-1}_{red} * J_{red})^{-1}
\end{equation}

The effective mass at the fingertip in response to a force applied along a direction $\hat{d}$ can then be modeled as:

\begin{equation}
    m_{eff} = \frac{1}{\hat{d}*\lambda_{tip}^{-1}*\hat{d}}
\end{equation}

This allows for the construction of the effective mass belted ellipsoids at the fingertip, where a vector connecting the center of the ellipsoid to a point on its circumference represents the effective mass as its magnitude and the direction it is felt in.

These ellipsoids in Fig.~\ref{m_eff}\,(a) reveal several characteristics of the Koala fingers that aid in grasping performance. Each ellipsoid is asymmetric, which can be traced to singularities inherent to the linkage mechanism and the underactuated degree of freedom. The singularities of the mechanism, represented by the large magnitude sections of the ellipsoids, are generally in-line with reaction forces from gripped objects, helping to increase grasp stability. The notable exception is the high mass in the negative z-direction at the beginning of the stroke, which allows the finger to execute pressing and pushing tasks. The underactuation of the finger affects $m_{eff}$ in the closing and opening directions, as shown in Fig.~\ref{m_eff}\,(b). $m_{eff}$ is low in the closing direction, as the underactuation complies to applied forces, while $m_{eff}$ is higher in the opening direction where the linkage is reduced to a 1-DoF system as the underactuation is grounded against a hard stop. This allows the gripper to safely adapt to the environment with low $m_{eff}$ without sacrificing stiffness while grasping. Ultimately, along the trajectory of the finger the effective mass felt at the finger tip with the RD actuator is on the order of tens of grams, demonstrating its ease of backdrivability.

\textbf{Force Parity}
The RD should be able to replicate the forces collected on the CD. To validate this, pinch grasps were executed on objects with thicknesses of 10\,mm to 40\,mm to compare the force-output of the CD and RD. The CD trigger was loaded under 100\,N of input force, the maximum human input it was designed for. Ten trials were conducted for each thickness, with the grasp force measured using a Biometrics P200 coin load cell. Results in Table \ref{Force_data} show that the RD successfully force matched the CD for all thicknesses.

\begin{table}[!ht]
    \centering
    \def\arraystretch{1.5}
    \caption{CD vs RD fingertip grasp forces}
    \begin{tabular}{c|cc|c}
        Thickness & CD & RD & Difference\\
        \hline
        10\,mm & $25.88 \pm 0.64 N$ &  $25.06 \pm 0.45 N$ & $-3.17\% \pm 4.25\%$\\
        20\,mm & $23.84 \pm 0.73 N$ & $24.88 \pm 0.91 N$ & $4.36\% \pm 6.73\%$ \\
        30\,mm & $24.41 \pm 0.65 N$ & $26.39 \pm 0.88 N$ & $8.11\% \pm 6.00\%$ \\
        40\,mm & $24.34 \pm 0.77 N$ & $26.81 \pm 1.16 N$ & $10.15\% \pm 7.50\%$ 
    \end{tabular}
    \label{Force_data}
\end{table}

\begin{figure*}[!t]
    \centering
    \includegraphics[width=\linewidth]{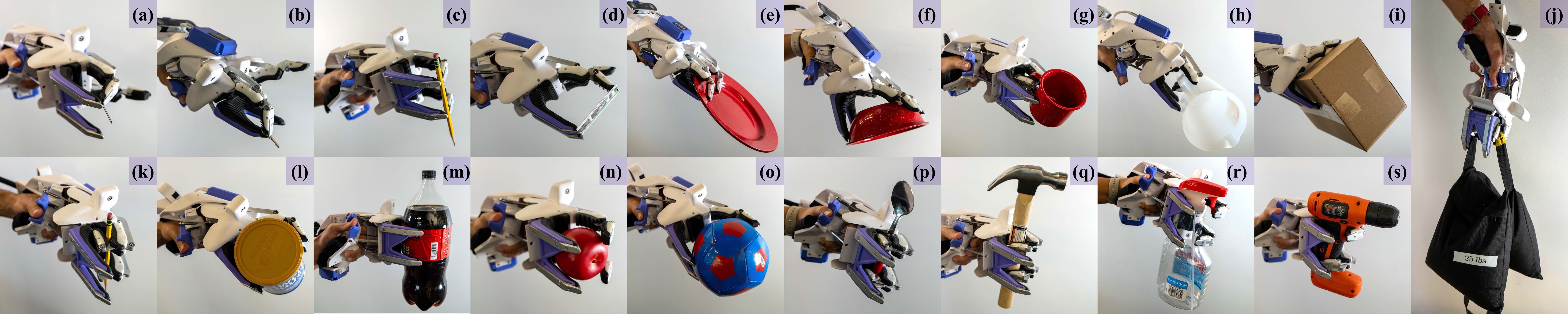}
    \caption{Koala grasps the following objects: (a) nail\textsuperscript{\dag}, (b) key\textsuperscript{\dag}, (c) pencil (pinch), (d) card, (e) plate\textsuperscript{\dag}, (f) bowl\textsuperscript{\dag}, (g) mug\textsuperscript{\dag}, (h) jug\textsuperscript{\dag}, (i) box, (j) 25\,lb sandbag, (k) pencil (power), (l) coffee can\textsuperscript{\dag}, (m) 2\,L bottle, (n) apple\textsuperscript{\dag}, (o) ball\textsuperscript{\dag}, (p) spoon\textsuperscript{\dag}, (q) hammer\textsuperscript{\dag}, (r) spray bottle\textsuperscript{\dag}, (s) drill\textsuperscript{\dag}. (\textsuperscript{\dag}\,From YCB object set \cite{Calli2015BenchmarkingSet})}   
    \label{grasps}
    \vspace{-3mm}
\end{figure*}

\subsection{Grasp Execution}

The Koala CD was used to grasp and manipulate a variety of objects and tools to qualitatively evaluate its capabilities (Fig.~\ref{grasps}). The gripper is able to grasp and use objects that are large and unwieldy (Fig.~\ref{grasps}\,(i,\,l,\,m,\,o)), have uneven weight distribution (Fig.~\ref{grasps}\,(g,\,q)), and are more complex functionality (Fig.~\ref{grasps}\,(r,\,s)). The fingernail features prove useful for singulating small objects down to nails with a pinch grasp, and allow the gripper to pick thin objects from flat surfaces, like a credit card in Fig.~\ref{grasps}\,(d). We additionally validate a maximum power grasp diameter of 115\,mm (Fig.~\ref{grasps}\,(m)) and minimum grasp diameter of 7\,mm (Fig.~\ref{grasps}\,(k)). Testing also shows that high force pinch grasps by the Koala gripper can emulate the effects of certain abducted thumb grasps like the key grasp (Fig.~\ref{grasps}\,(b)), extending the capabilities of the gripper beyond what its simplified morphology suggests.

\subsection{Policy Execution}

To validate that the data captured using the Koala platform is usable for imitation learning, we trained a diffusion-policy \cite{Chi2023DiffusionDiffusion} with a shared vision encoder across all camera streams and expanded the action and state dimensions to include the 3 DoF end-effector. For two example tasks, pasta straining and cup destacking, we collected 250 handheld and 100 teleoperated demonstrations respectively. 
Policies were rolled-out on a 7 DoF Franka Emika FR3 using impedance control for compliant execution.
Supplemental videos show representative successful completions, including behaviors enabled by the non-parallel jaw Koala gripper. 

%%%%%%%%%%%%%%%%%%%%%%%%%%%%%%%%%%%%%%%%%%%%%%%%%%%%%%%%%%%%%%%%%%%%%%%%
\section{Conclusion}

We have presented a co-design framework for paired data capture and robotic grippers that aims to improve platform capability by considering the constraints of both modalities from the beginning of the design process. We applied this design philosophy to the development of the Koala platform, which combined a 3-DoF morphology with unique finger and thumb mechanisms to improve on parallel jaw dexterity without compromising on ergonomics. The Koala gripper successfully collected data through both in-the-wild handheld collection and teleoperation, which was used to train diffusion policies to complete a series of tasks.

There are several areas for further development of this work. First, although the dual-thumb morphology recovers a wide range of grasps with only three controllable DoFs, certain grasps that benefit from thumb adduction (for example, in-hand re-orientation of small parts) remain out of reach. Adding an actuated adducting DoF on the thumb is the most direct route to closing this gap, at the cost of additional user mental load. Second, the full capabilities of the system are difficult to utilize in learned policies because of the lack of force-state information in collected data. Work is currently underway to add wrench and grip force sensing to the CD, which will enable force-reactive policies that can better harness the high force output of the Koala grippers. Further ergonomic development for the gripper would likewise be beneficial, as a limiting factor on the speed of data collection has been operator fatigue. Some avenues for improvement include moving the operator's hand closer to the active surfaces of the gripper, reducing torques on the operator's wrist, and conducting user studies to gain more comprehensive feedback on gripper fit and comfort. Finally, our policy demonstrations are limited to the tasks reported in Sec.~\ref{Testing}; characterizing how the embodiment gap between handheld and teleoperated supervision modalities scales with task complexity and dataset size is left to future work.

%%%%%%%%%%%%%%%%%%%%%%%%%%%%%%%%%%%%%%%%%%%%%%%%%%%%%%%%%%%%%%%%%%%%%%%%

\section*{Acknowledgments}

We thank Zachary Cecere, Peiyi Chen, Sandra Liu, Neehar Peri, Greg Xie, Jinxi Xu, Ge Zhang, and Yu Meng Zhou for their invaluable support working on this project.

%%%%%%%%%%%%%%%%%%%%%%%%%%%%%%%%%%%%%%%%%%%%%%%%%%%%%%%%%%%%%%%%%%%%%%%%

\bibliographystyle{IEEEtran}
\bibliography{references}

\end{document}